\documentclass[10pt,twocolumn,letterpaper]{article}

\usepackage{iccv}
\usepackage{times}
\usepackage{epsfig}
\usepackage{graphicx}
\usepackage{amsmath}
\usepackage{amssymb}
\usepackage{algpseudocode}
\usepackage{algorithm}
\usepackage{mathtools, nccmath}
\usepackage{soul}

\usepackage{caption}
\usepackage{subcaption}

\usepackage[symbol]{footmisc}

\DeclarePairedDelimiter{\nint}\lfloor\rceil

\usepackage[breaklinks=true,bookmarks=false]{hyperref}

\iccvfinalcopy 

\ificcvfinal\fi

\begin{document}

\title{Quantization Aware Factorization \\for Deep Neural Network Compression}

\author{\bf Daria Cherniuk, Stanislav Abukhovich, Anh-Huy Phan, \\
\bf Ivan Oseledets, Andrzej Cichocki, Julia Gusak\footnotemark{} \\
Skolkovo Institute of Science and Technology\\
{\tt\small \{daria.cherniuk, Stanislav.abukhovich, a.phan, i.oseledets, A.Cichocki \}@skoltech.ru} \\
{\tt\small julgusak@gmail.com}
}

\maketitle
\ificcvfinal\thispagestyle{empty}\fi

\begin{abstract}
   Tensor decomposition of convolutional and fully-connected layers is an effective way to reduce parameters and FLOP in neural networks. 
   Due to memory and power consumption limitations of mobile or embedded devices, the quantization step is usually necessary when pre-trained models are deployed. 
   A conventional post-training quantization approach applied to networks with decomposed weights yields a drop in accuracy. 
   This motivated us to develop an algorithm that finds tensor approximation directly with quantized factors and thus benefit from both compression techniques while keeping the prediction quality of the model. 
   Namely, we propose to use Alternating Direction Method of Multipliers (ADMM) for Canonical Polyadic (CP) decomposition with factors whose elements lie on a specified quantization grid.
   We compress neural network weights with a devised algorithm and evaluate it's prediction quality and performance.
   We compare our approach to state-of-the-art post-training quantization methods and demonstrate competitive results and high flexibility in achiving a desirable quality-performance tradeoff.
   
\end{abstract}

\section{Introduction}

\footnotetext[1]{Now at Inria, University of Bordeaux, France}

Recent years more and more applications based on deep learning algorithms appear on the market, and many of them are developed to work on mobile and edge devices. However, neural networks which are in the core of such algorithms can not usually be used as it is, because they either do not satisfy memory and energy constraints of devices or perform the inference phase not fast enough to satisfy the user needs. Thus, many approaches to reduce and accelerate neural networks have been proposed in the literature. 

Conventionally, these approaches include pruning, tensor/factorization methods, quantization, knowledge distillation, and architecture search.  
Thoughtful combination of different techniques gain their benefits simultaneously. 

The availability of additional information, such as training data or loss/gradients values, yields additional opportunities to improve performance of compressed models. From this point of view, we distinguish among techniques that do not use additional data, use data but do not perform training, and those that run several training epochs to tune parameters of the compressed model. 

In our paper we propose a method that jointly performs factorization and quantization of neural networks. Namely, we replace a convolutional layer represented with parameters in FLOAT32 format with a decomposed layer, which is constructed as a sequence of convolutional layers with weights in lower precision format (e.g., INT8/INT6/INT4). By applying our method we benefit from reduction in number of parameters and operations due to factorization and further model size reduction and acceleration due to lower bit representation of weights.  

Our main contributions are the following:

\begin{itemize}
    \item We introduce an ADMM-based algorithm, which approximates a tensor with a CP-factorization, where factors are represented in lower precision format, e.g., INT8/INT6/INT4. 
    \item We propose a neural network compression method that simultaneously performs factorization and quantization of weight tensors based on introduced tensor approximation technique. As far as we are concerned, this is the first work that introduces weights approximation by CP-decomposition with lower precision factors.
    \item We show that in terms of compression/accuracy trade-off our approach performs better than conventional tensor approximation methods followed by quantization of factors.
    \item We demonstrate the effectiveness of our method in comparison with other quantization approaches that are applicable when only a small set of training data is available. 
\end{itemize}

\section{Related Work}
There are many papers related to neural network speed-up and compression. In this section we review the most relevant to our proposed technique.

\subsection{Factorization}
Various tensor decompositions~\cite{Kolda2009, Cichocki2016} are used to factorize weights~\cite{panagakis2021tensor, kossaifi2019t, gusak2019automated} or activations~\cite{cui2020active, gusak2019reduced} of deep neural networks. These approaches allow to store less parameters during training and inference and to speedup the computation by reducing the number of required elementary operations. For linear layers, prior works have utilized Tensor-Train (TT) format ~\cite{Novikov2015}. For convolutional layers, whose weights are 4D tensors, authors in ~\cite{Lebedev2015} used CPD. Authors in ~\cite{Phan2020} also decomposed convolutional layers through CP-decomposition but proposed to flatten spatial dimension and factorize 3D tensors instead of the original 4D. Other tensor approximations techniques like Tucker~\cite{kim2015compression} or block term decompositions are used by researches as well. In our paper, we focus on CP decomposition for convolutional layers with kernel spatial dimensions greater than $1$ and matrix factorization for linear layers or 1x1 convolutional layers. We prefer to study these type of decompositions to others as they show promising accuracy/speed-up trade-off in existing research papers.

\subsection{Quantization}
Due to memory/power consumption limitations of edge devices, the quantization step is necessary, when pre-trained models are deployed. 
A comprehensive overview of papers on neural network quantization can be found in ~\cite{Gholami2022} and ~\cite{Nagel2021}.

Quantization might greatly impede the quality of a neural network prediction. 
There are different techniques to restore the quality, some of them requiring additional data or training.
Best accuracy restoration can be obtained with quantization-aware training: fine-tuning the quantized model while restricting model weights to be on the quantization grid. Another approach, taken in~\cite{BayesianBits}, is to learn bit-width and quantization parameters simultaneously with model weights. However, both techniques require dataset and resources to train.

Because of training expenses, methods that allow for more accurate quantization but avoid training/fine-tuning have been a research interest in several papers: authors in ~\cite{Eldad2019} scale the weights of consecutive layers which allows them achieve smaller quality drop on a wide variety of networks; AdaRound~\cite{nagel2020up} proves that rounding to the nearest node of the quantization grid is not the best strategy and propose to choose rounding by optimization process which requires only a few thousand samples of data, researches in ~\cite{Hubara2020} further develop the idea of AdaRound, complementing it with integer programming to determine the best bit-width for each layer.


\subsection{ADMM}
Alternating Direction Method of Multipliers or ADMM~\cite{Boyd2010} is an optimization algorithm for convex problems. The advantages of ADMM algorithm are that it can be efficiently parallelized~\cite{Shang2014, Liavas2015} and the convergence is guaranteed in convex case~\cite{Boyd2010}. However, for non-convex problems the algorithm is not guaranteed to converge or might not converge to a global minimum. Authors in ~\cite{Diamond2016} proposed using ADMM in conjunction with local improvement methods to find sufficient heuristic solutions.

In \cite{Huang2016} authors have applied ADMM to a task of finding a CP-decomposition that satisfied particular constraints like non-negativity, sparsity, etc.. They called this approach AO-ADMM, since one of the objectives was an Alternating Optimization(in particular, Alternating Least Squares) that is used to find CP factors. 

Authors in ~\cite{Yin2021} have embedded ADMM into training neural network to impose low TT-rank structure on its weights. After training, the weights are decomposed into a TT-format and fine-tuned. 

~\cite{Leng2018} aimed to apply ADMM to training neural network with weights that lie on a grid of scaled powers of two. The first objective function in their case is model prediction loss, the second - projection of weight tensors on the grid of scaled powers of two. The authors solve a discrete non-convex constraint problem by alternating between optimizing a scaling factor and projecting onto a fixed grid. 
The method converges, however, the projection even on the unscaled grid is a non-convex problem and is not guaranteed to converge to global minimum.

\section{Methodology}
\subsection{Convolutional Layer Factorization}\label{sec:conv_reshape}
Similar to \cite{Lebedev2015} and, later, \cite{Phan2020}, we represent convolution layer weights as a 3-way tensor by flattening spatial dimension and decompose it into 3 consecutive convolutional layers. Let $\boldsymbol{K} \in \mathbb{R}^{T \times S \times D \times D}$ be a kernel tensor, corresponding to convolutional layer with $S$ input channels, $T$ output channels, and $D \times D$ spatial convolution. Let $\overline{\boldsymbol{K}} \in \mathbb{R}^{T \times S \times D^2}$ denote the $\boldsymbol{K}$ tensor after reshape operation. Using rank-R CP decomposition, one element of $\overline{\boldsymbol{K}}$ can be represented as: 
\begin{equation}
    \overline{\boldsymbol{K}}(t, s, dd) \cong \sum^R_{r=1} \overline{K}^{dd}(dd,r)K^s(s, r)K^t(t,r)
\end{equation}
Then, in order to reshape convolutional layer back to original shape, we reshape factor-matrix $\overline{K}^{dd}$ to a tensor $\boldsymbol{K}^{dd}$ of shape $D \times D \times R$.
\begin{equation}
    \boldsymbol{K}(t, s, j, i) \cong \sum^R_{r=1} \boldsymbol{K}^{dd}(j,i,r)K^s(s,r)K^t(t,r)
\end{equation}
Therefore, having an input, $X$, to the convolutional layer, the output tensor, $Y$, is calculated as: 
\begin{equation}
\begin{split}
    & \boldsymbol{Y}(t, h', w') = \\
    & \sum_{h = h' - \delta}^{h'+\delta}\sum_{w = w' - \delta}^{w'+\delta}\sum_{s = 1}^S \boldsymbol{K}(t, s, h - h' + \delta, w - w' + \delta) \boldsymbol{X}(s, h, w)  \\
\end{split}
\end{equation}
Substituting kernel expression into the formula above, performing rearrangements and grouping summands, we obtain the following consecutive expressions for the approximate evaluation of the convolution
\begin{align*}
    & \boldsymbol{Z}^1(r, h, w) = \sum_{s = 1}^S K^s(s, r)\boldsymbol{X}(s, h, w),\\
    & \boldsymbol{Z}^2(r, h', w') = \\
    & \sum_{h = h' - \delta}^{h'+\delta}\sum_{w = w' - \delta}^{w'+\delta} \boldsymbol{K}^{dd}(h - h' + \delta, w - w' + \delta, r)\boldsymbol{Z}^1(r, h, w),\\
    & \boldsymbol{Y}(t, h', w') = \sum_{r = 1}^R K^t(t, r)\boldsymbol{Z}^2(r, h', w'),
\end{align*}
where $\delta = D/2$. 

Therefore, we substitute original convolution with a sequence of smaller convolutions: point-wise convolution that reduces the number of input channels from $S$ to $R$, convolution with the same spatial dimension as the original but with $R$ number of  input and output channels, and another point-wise convolution that changes the number of channels from $R$ to $T$. The last convolution in sequence adds the original bias. 

\paragraph{1x1 kernel size} Convolutional layer with 1x1 kernel is equivalent to a Linear layer with weight matrix of shape $S \times T$ acting on the input with flattened spatial dimension and, thus, can be decomposed into two matrix factors and replaced with a sequence of two convolutional layers both with kernel size 1x1.


\subsection{Quantization}
Quantizing of neural network means transforming its weights and/or activations into low-bit fixed-point representations, e.g., INT8. It saves memory occupied by model parameters, reduces the amount of data transfer, size and energy consumption of the MAC operation ~\cite{Nagel2021}.

In our work we use both signed symmetric and asymmetric uniform per-tensor quantization ~\cite{Nagel2021} in which tensors are mapped to their int versions in the following way:
\begin{equation}
\begin{split}
    & x_{\text{int}} = \text{clip} \left(\nint{\frac{x}{\mathrm{scale}}} + z; -2^{b-1}, 2^{b-1} -1\right),
\end{split}
\end{equation}
where $\nint{x}$ denotes rounding $x$ to the nearest integer, $\mathrm{scale}$ is a step of the quantization grid, $z$ is a zero-point and $b$ is a number of bits in quantization. In case of symmetric quantization $z$ is assumed to be $0$. 
Real-valued approximation is obtained by extracting zero-point and multiplying by scale: $x \approx \mathrm{scale} * (x_{\text{int}} - z)$.
For per-tensor MinMax quantization ~\cite{Nagel2021}, $\mathrm{scale}$ is determined from minimum and maximum values of tensor $X$:
\begin{equation}
\begin{split}\label{eq:quantization}
    & \mathrm{scale} = \frac{\max\{X\} - \min\{X\}}{2^b-1} \\
\end{split}
\end{equation}
MinMax quantization, however, might suffer from large outliers. A way to alleviate this issue is to use MSE approach to set range of values~\cite{Nagel2021}. In case of symmetric quantization (to reduce number of optimized parameters from 2 to 1 and provided that the value distribution is symmetric, as shown in Figure \ref{fig:factor_hists}), $\mathrm{scale}$ calculation takes the following form:
\begin{equation}
\begin{split}
    & \mathrm{scale} = \frac{2q_{\max}}{2^b-1}
\end{split}
\end{equation}
and $q_{\max}$ is determined through optimizing 
\begin{equation}
    \underset{q_{\max}}{\arg\min} \lVert X - \hat{X}(q_{\max}) \rVert^2_F
\end{equation}
where $\hat{X}$ is a quantized tensor $X$.
In Section ~\ref{sec:experiments}, we perform an ablation study to determine the best quantization scheme (Figure~\ref{tab:minmaxvsmse}). 

\subsection{Quantization-aware factorization}
Finding a decomposition that produces less error after quantization can be formulated as a constrained tensor factorization problem. A method called AO-ADMM(a hybrid of the Alternating Optimization and the Alternating Direction Method of Multipliers) has been shown~\cite{Huang2016} to successfully solve problems that involve  non-negativity, sparsity or simplex constraints. In our work we extend the field of its application by introducing a constraint function that ensures low quantization error of the derived factors and construct a corresponding algorithm. 

Searching for a factorization of $n \times m$ matrix $X$ that satisfies quantization constraint (i.e. obtained factors are equal to their quantized versions) means minimizing the following function:
\begin{equation}
\begin{split}\label{eq:task}
    & \underset{A,B}{\text{minimize }}
    \frac{1}{2} \lVert X - AB^T \rVert^2_F + I_Q(A) + I_Q(B) \\
\end{split}
\end{equation}
where $A$ is a matrix of shape $n \times r$, $B$ - of shape $r \times m$ ($r \leq n,m$) and $I_Q$ is an indicator function such that $I_Q(Y)=0$ when $Y \in Q$ and $I_Q(Y)=+\infty$ if $Y \notin Q$. In case of symmetric per-tensor quantization ~\cite{Nagel2021}, $Q$ is a set of tensors whose elements belong to a discrete quantization grid $\{ (-2^{b-1}) * \mathrm{scale}, ..., (2^{b-1} - 1) * \mathrm{scale}\}$.

We introduce an auxilary variable, $\tilde{B}$, and formulate a constrained Least Squares problem, which can be solved using alternating update scheme by fixing one variable and minimizing the objective function over the other. For fixed factor, $A$, the subproblem takes the following form: 
\begin{equation}
\begin{split}\label{eq:task_b}
    & \underset{B,\tilde{B}}{\text{minimize }}
    \frac{1}{2} \lVert X - A\tilde{B} \rVert^2_F + I_Q(B) \\
    & \text{subject to } B = \tilde{B}^T
\end{split}
\end{equation}

The ADMM method introduces a dual variable, $U$, for the equality constraint, $B = \tilde{B}^T$, and alternates between optimizing each part of the objective function:

\begin{align}\label{eq:admm}
    & \tilde{B} = \arg \underset{\tilde{B}}{\min} \left( \frac{1}{2} \lVert X - A \tilde{B} \rVert^2_F + \frac{\rho}{2} \lVert B - \tilde{B}^T + U \rVert^2_F \right) \\
    & B = \arg\underset{B}{\min} \left(I_Q(B)+ \frac{\rho}{2} \lVert B - \tilde{B}^T + U \rVert^2_F \right) \\
    & U = U + B - \tilde{B}^T
\end{align}

The optimal solution for the auxiliary variable $\tilde{B}$ is given in closed form:

\begin{equation}
\begin{split}\label{eq:b_tilda}
    & \tilde{B} = (A^TA + \rho I)^{-1} (X^T A + \rho(B + U))^T \\
\end{split}
\end{equation}

For efficiency, (\ref{eq:b_tilda}) is solved through Cholesky decomposition of $A^TA + \rho I$.

Solving (\ref{eq:admm}) for the factor, $B$, gives the minimum distance between $(\tilde{B}^T - U)$ and set $Q$:

\begin{equation}
    \begin{split}
        B &= \text{arg} \underset{B \in Q}{\text{min }} \lVert B - \tilde{B}^T + U \rVert^2_F  \\
          &= \mathrm{proj}_Q (\tilde{B}^T - U ) \label{eq:proj}
    \end{split}
\end{equation}

\begin{algorithm}
	\caption{Solve \ref{eq:task_b} using ADMM} \label{alg:admm}
	 \hspace*{\algorithmicindent} \textbf{Input:} B, U ,K, G, $\mathrm{rank}$ \\
    \hspace*{\algorithmicindent} \textbf{Output:} B, U 
	\begin{algorithmic}[1]
		\State $\rho \gets \mathrm{trace}(G) / \mathrm{rank}$
		\State $L \gets \mathrm{Cholesky}(G + \rho I)$
		\Repeat
		    \State $\tilde{B} \gets L^{-T}L^{-1} (K + \rho(B+U))^T$
		    \State $B_0 \gets B$
		    \State $B \gets \mathrm{proj}_Q(\tilde{B}^T - U)$
		    \State $U \gets U + B - \tilde{B}^T$
		    \State $r \gets \lVert B-\tilde{B}^T \rVert^2_F / \lVert B \rVert^2_F $
		    \State $s \gets \lVert B - B_0 \rVert^2_F / \lVert U \rVert^2_F $
		\Until{ $r < \epsilon$ and $s < \epsilon$}
	\end{algorithmic} 
\end{algorithm}

Overall, the ADMM procedure for the factor, $B$, is defined in Algorithm~\ref{alg:admm}. For $X \approx A\tilde{B}^T$ matrix factorization $G$ is a Gram matrix of the factor that is fixed on current iteration of Alternating 
Least Squares and $K$ is an original matrix multiplied by a fixed factor ($A^TA$ and $X^TA$ from (\ref{eq:b_tilda})). We adopt expressions for $r$, $s$ and $\rho$ from ~\cite{Huang2016}. Similarly, the task is solved for the remaining factor through minimizing  $|| X - \tilde{A}^TB^T||_F^2$ over set $Q$. The procedure continues alternating between factors until convergence.

For 3-way tensors, the matrix multiplication of two factors in (\ref{eq:task}) is replaced with the Canonical Polyadic decomposition (CP):
\begin{equation}
    \begin{split}\label{eq:task_3d}
        \underset{A,B,C}{\text{minimize}} (
        \frac{1}{2} \lVert \mathbf{X} - \sum^R_{r=1} A_{:,r} \otimes B_{:,r} \otimes C_{:,r}\rVert^2_F & \\
         + I_Q(A) + I_Q(B) +  I_Q(C) ) &  \\
    \end{split}
\end{equation}
where $\mathbf{X}$ is a tensor of shape $n \times m \times k$, A is a matrix of shape $n \times r$, B - matrix of shape $m \times r$, C - matrix of shape $k \times r$, $\otimes$ denotes the outer product. 
The constrained ALS subproblem for factor $B$ takes the following form:
\begin{equation}
    \begin{split}\label{eq:task_3d_b}
         & \underset{B,\tilde{B}}{\text{minimize }}
        \frac{1}{2} \lVert \mathbf{X}_{(2)} - \tilde{B}^T(C \odot A)^T \rVert^2_F + I_Q(B) \\
        & \text{subject to } B = \tilde{B}^T
    \end{split}
\end{equation}
where $\mathbf{X}_{(2)}$ is a matrix unfolding of tensor $\mathbf{X}$ by mode $2$, $\odot$ denotes Khatri-Rao product. The ADMM procedure is the same as in Algorithm~\ref{alg:admm}, with $G = (A^TA) * (C^TC)$ and $K = \mathbf{X}_{(2)}(C \odot A)$ ~\cite{Smith2017}. Optimization with respect to factors $A$ and $C$ is done in the same manner ~\cite{Smith2017}. Alternating between factors proceeds until convergence. 

\begin{algorithm}
	\caption{Solve (\ref{eq:task_3d}) using ADMM-EPC} \label{alg:admm}
	 \hspace*{\algorithmicindent} \textbf{Input:} $\mathbf{X}$, $\mathrm{rank}$ \\
    \hspace*{\algorithmicindent} \textbf{Output:} A, B, C
	\begin{algorithmic}[1]
		\State Initialize: $A$, $B$, $C$ with CPD-EPC 
		\State Initialize: $U_A$, $U_B$, $U_C$ with $0$ 
		\Repeat
		    \State $G \gets B^TB * C^TC$
            \State $K \gets \mathbf{X}_{(1)} (C \odot B)$
            \State $A, U_A \gets \mathrm{ADMM} (A, U_A, K, G, \mathrm{rank})$
            \\
		    \State $G \gets A^TA * C^TC$
            \State $K \gets \mathbf{X}_{(2)} (C \odot A)$
            \State $B, U_B \gets \mathrm{ADMM} (B, U_B, K, G, \mathrm{rank})$
            \\
            \State $G \gets A^TA * B^TB$
            \State $K \gets \mathbf{X}_{(3)} (B \odot A)$
            \State $C, U_C \gets \mathrm{ADMM} (C, U_C, K, G, \mathrm{rank})$
		\Until{ $e_{\mathrm{quant}}$(\ref{eq:quant_rec_err}) ceases to improve}
	\end{algorithmic} 
\end{algorithm}

\begin{figure}[t]
\begin{center}
    \includegraphics[width=0.9\linewidth]{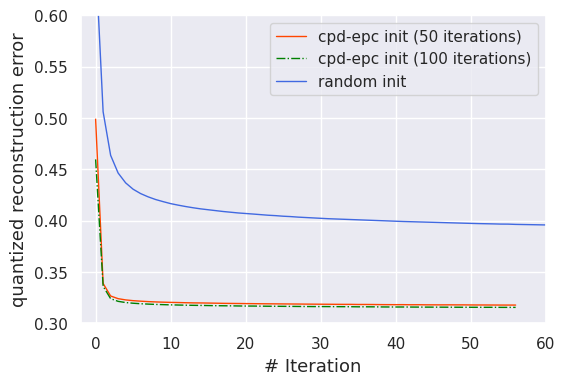}
\end{center}
\caption{Convergence of ADMM algorithm for a single layer of ResNet18 model with random and CPD-EPC initialization. We have found that using CPD-EPC as initialization instead of random sampling leads to lower quantized reconstruction error, while only a few number of CPD-EPC iterations is sufficient. This result reproduces throughout all model layers.}
\label{fig:quant_rec_error_init}
\end{figure}

The set $Q$ is non-convex, thus, it is not guaranteed that solution converges to global or even local minimum. Moreover, as shown in Figure~\ref{fig:quant_rec_error_init}, convergence depends on initialization. Authors in ~\cite{Diamond2016} proposed to use Multistart, Polishing and Neighbor search to improve solution. However, in our case, search among neighbors(even sampling search) of current iteration factor would require number of computations comparable to the number of parameters in a layer. Polishing on $Q$ reduces to simple recomputation of $\tilde{B}_{k+1}$ with $B_{k+1}$ instead of $B_k$ since the only convex restriction~\cite{Diamond2016} that can be chosen in this case is the trivial one. Carefully choosing initialization, however, proved to be a plausible technique to obtain a better solution (Figure ~\ref{fig:quant_rec_error_init}).

\section{Experiments}\label{sec:experiments}

We use pretrained ResNet18 model from torchvision model zoo\footnote{https://pytorch.org/vision/stable/models.html} and ImageNet\footnote{https://www.image-net.org/} dataset to evaluate our method. 
First, we conduct several ablation studies to determine quantization scheme, initialization, convergence properties e.t.c..
Then, we compare variations of our method with state-of-the-art approaches from previous papers as well as a naive approach of directly quantizing factors.
In all our experiments we utilize per-tensor quantization. 
For fair comparison, we only consider methods that do not perform model fine-tuning and require only a small sub-sample of the training dataset.

\paragraph{Metrics}
In some ablation studies we compute a quantized reconstruction error metric: 
\begin{equation}\label{eq:quant_rec_err}
    e_{\mathrm{quant}} = \frac{\lVert \mathbf{X}_{(2)} - \hat{B}(\hat{C} \odot \hat{A})^T \rVert_F}{\lVert \mathbf{X}_{(2)} \rVert_F}
\end{equation}
where $\hat{A}$, $\hat{B}$, $\hat{C}$ are quantized factors of tensor $\mathbf{X}$. $\mathbf{X}_{(2)}$ is a matrix unfolding by mode $2$. We could use any other unfolding, the purpose of this notation is merely to address the fact that Frobenius norm is formulated for matrices. 

Taking into consideration benefits of reducing both MAC operations (factorization) and operand's bit-width (quantization), we adopt BOP metric, introduced in~\cite{BayesianBits} and use it to compare our method with other approaches. For layer $l$, BOP count is computed as follows:
\begin{equation}\label{eq:bops}
    BOPs(l) = MACs(l)b_w b_a
\end{equation}
where $b_w$ is a bit-width of weights and $b_a$ is a bit-width of (input) activations. While computing this metric for the whole model we take into consideration all layers including BatchNorms. 

\paragraph{Factorization ranks} 
To choose a rank for each layer factorization, we set up a parameter reduction rate (i.e. the ratio of the number of the factorized layer's parameters over that of the original layer) and define the rank as 
$N/(n + m)/rate$ for fully-connected layers and 
$N/(n + m + k)/rate$ for convolutions. 
$N$ is the number of parameters in the original layer, $n$, $m$ and $k$ - dimensions of reshaped convolution weights (Section~\ref{sec:conv_reshape} ), $rate$ denotes parameter reduction rate. 

\paragraph{BatchNorm calibration}
Since Resnet models have a lot of BatchNorm layers, and both factorization and quantization disturb the distribution of outputs, calibrating BatchNorm layers' parameters (performing inference with no gradients or activations accumulation) can boost model prediction quality. 
The same procedure was used, for example, in AdaQuant\cite{Hubara2020}.
In all our experiments we perform calibration on $2048$ samples from training dataset. It is the same number of samples as used in AdaRound\cite{nagel2020up} and AdaQuant.

\subsection{Ablation studies}\label{seq:ablation}

\begin{figure}[t]
\begin{center}
\begin{subfigure}[b]{0.5\textwidth}
  \centering
  \includegraphics[width=.9\linewidth]{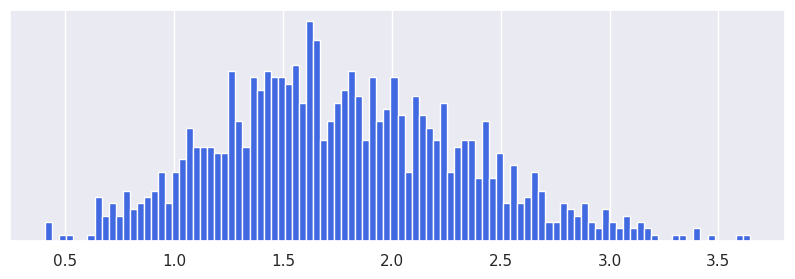}
  \caption{CPD}
  \label{fig:cpd_hist}
\end{subfigure}
\begin{subfigure}[b]{0.5\textwidth}
  \centering
  \includegraphics[width=.9\linewidth]{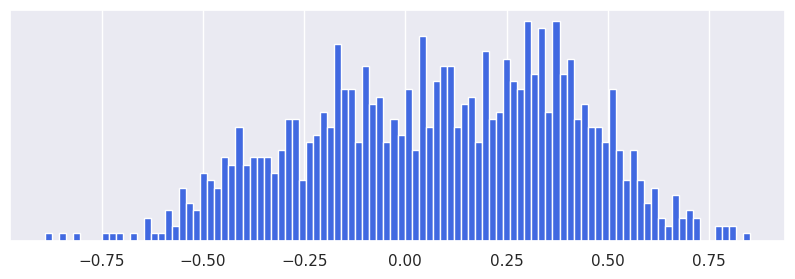}
  \caption{CPD-EPC}
  \label{fig:cpd_epc_hist}
\end{subfigure}
\begin{subfigure}[b]{0.5\textwidth}
  \centering
  \includegraphics[width=.9\linewidth]{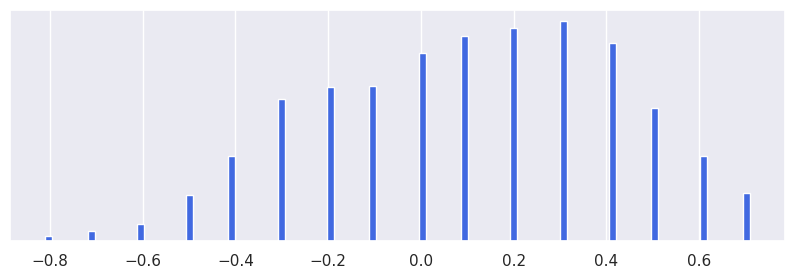}
  \caption{ADMM with CPD-EPC initialization}
  \label{fig:admm_epc_hist}
\end{subfigure}
\begin{subfigure}[b]{0.5\textwidth}
  \centering
  \includegraphics[width=.9\linewidth]{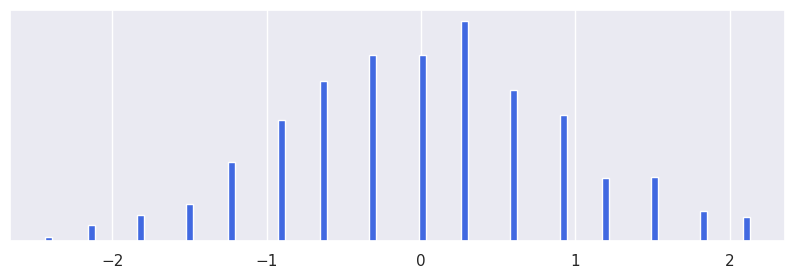}
  \caption{ADMM with random initialization}
  \label{fig:admm_hist}
\end{subfigure}
\end{center}
\caption{Histograms of values of factor matrix obtained with different decomposition methods from a single layer (layer1.0.conv1) of the ResNet18 model. Our ADMM-based algorithm produces factors whose values lie on a specified quantization grid.}
\label{fig:factor_hists}
\end{figure}

\paragraph{CP factors quantization}
First, we try to perform factorization and quantization successively. We obtain factors using CPD and CPD-EPC~\cite{Phan2020} and compare results after quantizing weights while leaving activations in full precision (Table~\ref{tab:cpdvsepc}).
Although CPD factors provide close to unfactorized model accuracy, they degrade to very poor quality of predictions upon quantization. 
CPD-EPC factors provide better model accuracy after quantization, while having lower accuracy before quantization. This is explained by the fact that CPD-EPC minimizes Sensitivity~\cite{sensitivity} of a decomposition that results in factors with lower value range then that of the classic CPD (Figure~\ref{fig:factor_hists}). It alleviates the problem of large outliers for uniform quantization.

We have also found (Figure~\ref{fig:quant_rec_error_init}) that using a few iterations of CPD-EPC as initialization of ADMM algorithm decreases quantized reconstruction error (\ref{eq:quant_rec_err}) by keeping value distribution narrowly around zero (Figure~\ref{fig:factor_hists}) which ensures a better quality of model predictions after quantization, especially when using symmetric quantization schemes. 
We have observed that only a few iterations of cpd-epc is sufficient, after which accuracy of a quantized model ceases to improve.

\begin{table}
\begin{center}
\begin{tabular}{|l|c|c|}
\hline
\#bits W/A & CPD & CPD-EPC \\
\hline\hline
32/32 & 69.26 & 66.65 \\
8/32 & 0.1 & 66.5 \\
6*/32 & 0.1 & 65.47 \\
4*/32 & 0.1 &  37.54 \\
\hline
\end{tabular}
\end{center}
\caption{Comparison (\% top-1 test accuracy) between CPD and CPD-EPC decompositions of a ResNet18 model(with reduction rate 2) before and after per-tensor MinMax quantization of weights. *We quantize all factorized layers to a specified bit-width, but leave batchnorm, downsample, first and last layers in $8$ bits.}\label{tab:cpdvsepc}
\end{table}

\paragraph{Quantization scheme}
We conducted an ablation study to choose between MinMax and MseMinMax quantization schemes to be used in a projection step of ADMM algorithm (\ref{eq:proj}). 
Results are presented in Table~\ref{tab:minmaxvsmse}. 
Although MSEMinMax slightly increases the overall computation time, it produces consistently better results due to scale adjustment. 
Hence, we use MSEMinMax scheme in all subsequent experiments.

\begin{table}
\begin{center}
\begin{tabular}{|l|c|c|}
\hline
\#bits W/A & MinMax & MseMinMax \\
\hline\hline
8/8 & 67.99 & 68.56 \\
6*/8 & 67.51 & 68.41 \\
4*/8 & 62.66 &  67.75 \\
\hline
\end{tabular}
\end{center}
\caption{Comparison (\% top-1 test accuracy)  between per-tensor MinMax and MSEMinMax quantization schemes used in projection step of ADMM iteration (the algorithm was run with reduction rate $2$). Post-factorization quantization of weights is performed with per-tensor MSEMinMax scheme in both cases. Quantization of activations is performed by histogram observer evaluated on 500 samples. *We quantize all factorized layers to a specified bit-width, but leave batchnorm, downsample, first and last layers in $8$ bits.}\label{tab:minmaxvsmse}
\end{table}

\begin{figure}
     \centering
     \begin{subfigure}[b]{0.5\textwidth}
         \centering
         \includegraphics[width=\textwidth]{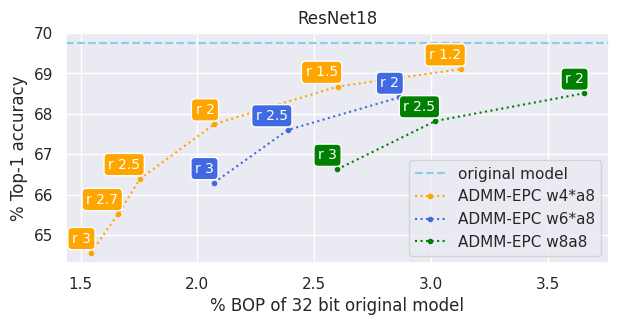}
         \caption{ADMM}
         \label{fig:resnet18_admm}
     \end{subfigure}
     \hfill
     \begin{subfigure}[b]{0.5\textwidth}
         \centering
         \includegraphics[width=\textwidth]{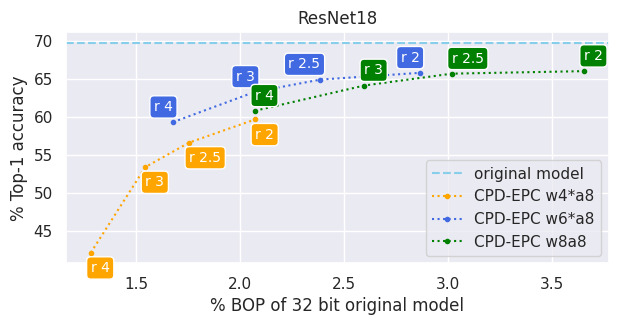}
         \caption{CPD-EPC}
         \label{fig:resnet18_epc}
     \end{subfigure}
        \caption{Ablation studies for best bit-width configuration. wXaY denotes quantization of weights with bit-width X and activations with bit-width Y. rZ labels indicate that factorization was performed with reduction rate Z. ADMM-EPC defines our ADMM-based approach with CPD-EPC initialization. *We quantize all factorized layers to a specified bit-width, but leave batchnorm, downsample, first and last layers in $8$ bits.}
        \label{fig:three graphs}
\end{figure}

\paragraph{Bit-allocation}
Determining the best trade-off between bit-allocation and reduction rate is not a trivial task. We performed several ablation studies (Figures ~\ref{fig:resnet18_admm} and ~\ref{fig:resnet18_epc}) to choose between different configurations and came to the conclusion that our ADMM-based approach benefits most from lower bit-width, while for conventional decomposition followed by quantization, the best results were achieved with 6-bit weight quantization.

\begin{figure*}
\begin{center}
    \includegraphics[width=\linewidth]{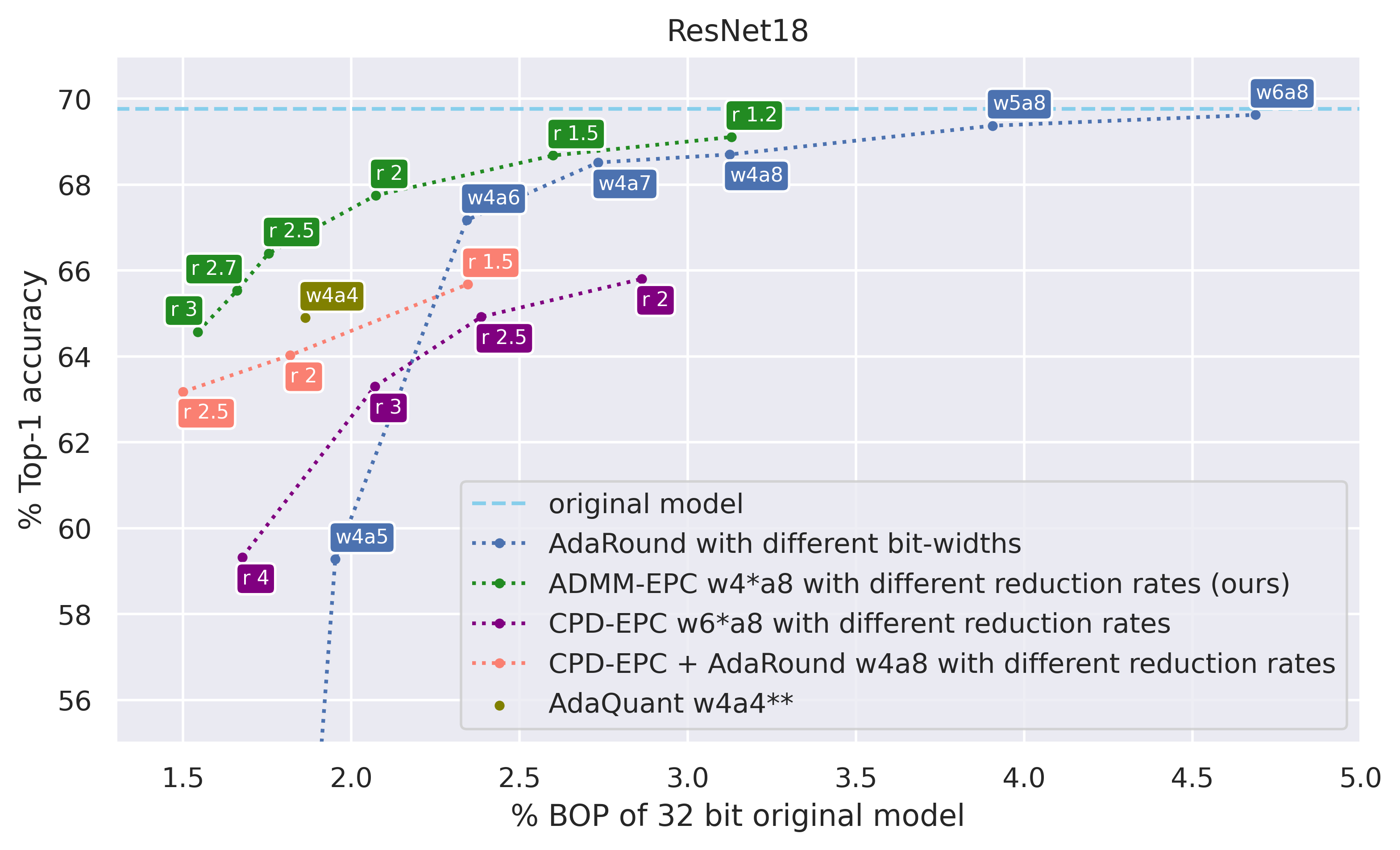}
\end{center}
   \caption{Comparison between post-training quantization methods on ResNet18 model. wXaY denotes quantization of weights with bit-width X and activations with bit-width Y. rZ labels indicate that factorization was performed with reduction rate Z. ADMM-EPC defines our ADMM-based approach with CPD-EPC initialization. *We quantize all factorized layers to a specified bit-width, but leave batchnorm, downsample, first and last layers in $8$ bits. **AdaQuant quantizes all layers to $4$ bits, except for the first and last layer which are quantized to $8$ bits. AdaQuant uses per-channel quantization.}
\label{fig:resnet18}
\end{figure*}

\subsection{Baselines}\label{seq:baselines}

We use top-1 test accuracy and BOP count (\ref{eq:bops}) to compare our method to other post-quantization approaches: AdaRound\cite{nagel2020up}, AdaQuant\cite{Hubara2020}, AdaRound applied to factors obtained with CPD-EPC decomposition and naive approach of successive CPD-EPC factorization and quantization. 

We used AdaRound implementation from the AIMET\footnote{https://github.com/quic/aimet} framework and AdaQuant from the author's repository\footnote{https://github.com/itayhubara/CalibTIP}. AdaRound uses per-tensor quantization, while AdaQuant performes per-channel quantization of weights. 
For AdaQuant, we used the advanced pipeline \cite{Hubara2020} with bias-tuning, even though it involves quantization-aware training step. We report the best accuracy we could achieve, though it is not equal to accuracy depicted in AdaQuant paper. 
But considering $71.97\%$ FP32 benchmark reported for ResNet18 model (in comparison with $69.76\%$ used as starting point in this paper), the quality drop remains the almost the same. We have also noticed that AdaQuant does not consider quantizing inputs to residual connection summation. 

With ADMM, CPD-EPC and CPD-EPC+AdaRound methods we do not factorize the first, last and downsample layers of the model. 
These layers contain small number of parameters but are placed in crucial parts of the architecture. 
Thus, factorizing them will not bring sufficient reduction in mac operations but might significantly degrade accuracy of predictions.
Due to the fact that we do not factorize these layers, their weights still lie in full range of FP32 values in contrast with ADMM-factorized layers whose values lie on a uniform quantization grid of a specific bit-width. 
We leave these layers to 8-bit quantization for ADMM and CPD-EPC benchmarks. 
This distribution of bit-widths is concordant with other works on quantizing resnet-like architectures, e.g., \cite{Hubara2020}, even though factorization of layers is not involved. 

Results are presented in Figure~\ref{fig:resnet18}, more details can be found in the supplementary materials.
It can be seen that our ADMM-based method outperforms other approaches on smaller BOP count values and shows competitive results on larger ranges. 

Another benefit of a joint factorization-quantization approach is that it is more adaptive in achieving a desirable trade-off between accuracy and efficiency: varying reduction rates gives more flexibility than changing bit-widths configurations.

\section{Conclusion}
\label{sec:conclusion}
In this paper we proposed a technique for compressing neural networks that performs joint factorization and quantization. We introduced ADMM-based approximation algorithm to replace weights with float32 elements with their factorized representations, whose factors contain elements of lower precision (int8/int6/int4). 

Our method allows to benefit from both factorization and quantization techniques, resulting in decrease of model size and acceleration in model inference. 
Our experiments have shown that ADMM-based joint quantization and factorization shows superior or on par results in comparison to other recent approaches to post-training quantization that do not involve model training and only require a small subset of training data. 
Moreover, our method allows for a flexible trade-off between accuracy and acceleration not restricted solely by bit-width configurations. 

Nevertheless, there are a lot of opportunities to improve our method. For example, incorporating more sophisticated tensor and quantization techniques.
Another possible direction would be to devise an algorithm of jointly choosing best factorization rank and bit-width similar Bayesian optimization in PARS~\cite{Sobolev2022} or integer programming approach in ~\cite{Hubara2020}.

{\small
\bibliographystyle{ieee_fullname}
\bibliography{bib/daria,bib/julia}
}

\end{document}